\documentclass[conference]{IEEEtran}
\usepackage{cite}
\usepackage{caption}
\ifCLASSINFOpdf
   \usepackage[pdftex]{graphicx}
\else
\fi
\usepackage[cmex10]{amsmath}
\usepackage{algorithmic}
\usepackage{algorithm}
\usepackage{verbatim}
\usepackage{amssymb}

\usepackage{array}
\usepackage{mathtools}

\usepackage{url}
\usepackage{booktabs}
\usepackage{multirow}
\usepackage{tabularx}
\usepackage[table]{xcolor}
\newcolumntype{C}[1]{>{\raggedleft\let\newline\\\arraybackslash\hspace{0pt}}m{#1}}

\newcommand\R{{\mathbb R}}

\usepackage{multirow}
\usepackage{blindtext}
\usepackage{algorithm}
\usepackage{algorithmic}
\usepackage{multirow}
\usepackage{tabularx}
\usepackage[table]{xcolor}
\usepackage{blindtext}
\usepackage{graphicx}
\usepackage{verbatim}
\usepackage{amssymb}

\usepackage{amsmath}
\DeclareMathOperator*\argmax{\operatorname{argmax}}

\begin{document}
\title{Time-Series Classification Through \\ Histograms of Symbolic Polynomials }


\author{\IEEEauthorblockN{Josif Grabocka, Martin Wistuba, Lars Schmidt-Thieme}
\IEEEauthorblockA{Information Systems and Machine Learning Lab\\
University of Hildesheim\\
\{josif, wistuba, schmidt-thieme\}@ismll.uni-hildesheim.de }}


%Telephone: (800) 555--1212, Fax: (888) 555--1212}


\maketitle

\begin{abstract}
Time-series classification has attracted considerable research attention due to the various domains where time-series data are observed, ranging from medicine to econometrics. Traditionally, the focus of time-series classification has been on short time-series data composed of a unique pattern with intra-class pattern distortions and variations, while recently there have been attempts to focus on longer series composed of various local patterns. This study presents a novel method which can detect local patterns in long time-series via fitting local polynomial functions of arbitrary degrees. The coefficients of the polynomial functions are converted to symbolic words via equivolume discretizations of the coefficients' distributions. The symbolic polynomial words enable the detection of similar local patterns by assigning the same words to similar polynomials. Moreover, a histogram of the frequencies of the words is constructed from each time-series' bag of words. Each row of the histogram enables a new representation for the series and symbolize the existence of local patterns and their frequencies. Experimental evidence demonstrates outstanding results of our method compared to the state-of-art baselines, by exhibiting the best classification accuracies in all the datasets and having statistically significant improvements in the absolute majority of experiments.
\end{abstract}



% \fi
\IEEEpeerreviewmaketitle

\section{Introduction}

Classification of time series is an important domain of machine learning due to the widespread occurrence of time-series data in real-life applications. Measurements conducted in time are frequently encountered in diverse domains ranging from medicine and econometrics up to astronomy. Therefore, time series has attracted considerable research interest in the last decade and a myriad of classification methods have been introduced. 

Most of the existing literature on time-series classification focuses on classifying short time series, that is series which mainly incorporate a single long pattern. The research problem within this family of time-series data is the detection of pattern distortions and other types of intra-class pattern variations. Among other successful techniques in this category, the nearest neighbor classifier equipped with a similarity metric called Dynamic Time Warping (DTW) has been shown to perform well in a large number of datasets\cite{Ding:2008:QMT:1454159.1454226}.

Nevertheless, few studies 
\cite{DBLP:conf/gfkl/BuzaS08,Lin:2009:FSS:1561638.1561679,DBLP:journals/jiis/0001KL12} have been dedicated towards the classification of time-series data which is long and composed of many local patterns. Avoiding the debate on the categorization of the words \emph{short} and \emph{long}, we would like to emphasize the differences between those two types of data with a clarification analogy. For instance, assume we would like to measure the similarity between two strings (i.e. words or time-series of characters). In such a case, string similarity metrics will scan the characters of those two strings, in a sequence, and detect the degree of matchings/similarities. In contrast, if we are searching for similarities between two books, then a character by character similarity is meant to fail because the arrangement of words in different books is never exactly the same. Text mining approaches which compute bags of words occurring in a text and then utilize histograms of the occurrences of each word, have been recently applied to the time-series domain by the metaphor of computing "bags of patterns" \cite{DBLP:journals/jiis/0001KL12,DBLP:conf/gfkl/BuzaS08}. 

This paper presents a novel method to classify long time series composed of local patterns occurring in an unordered fashion and by varying frequencies. Our principle relies on detecting local polynomial patterns which are extracted in a sliding window approach, hence fitting one polynomial to each sliding window segment. As will be detailed in Section~\ref{fitpolySec} we propose a quick technique to fit sliding window content which has a linear run-time complexity.

Once the polynomial coefficients of each sliding window content are computed, then we convert those coefficients into symbolic forms, (i.e. alphabet words). The motivation for calling the method Symbolic Polynomial arises from that procedure. Such discretization of polynomial coefficients, in the form of words, allows the detection of similar patterns by converting close coefficient values into the same literal word. In addition, the words computed from the time series allow the construction of a dictionary and a histogram of word frequencies, which enables an efficient representation of local patterns.

We utilize an equivolume discretization of the distributions of the polynomial coefficients to compute the symbolic words, as will be explained in Section~\ref{wordsSec}. Threshold values are computed to separate the distribution into equal volumes and each volume is assigned one alphabet letter. Consequently, each polynomial coefficient is assigned to the region its value belongs to, and is replaced by the region's character. Ultimately, the word of a polynomial is the concatenation of the characters of each polynomial coefficient merged together. The words of each time series are then stored in a separate 'bag'. A dictionary is constructed with each word appearing at least once in the dataset and a histogram is initialized with each row representing a time series and each column one of the words in the dictionary. Finally, the respective frequencies of words are updated for each time series and the rows of the histogram are the new representation of the original time series. Such a representation offers a powerful mean to reflect which patterns (i.e. symbolic polynomial words) and how often they occur in a series (i.e. the frequency value in each histogram cell).

The novelty of our method, compared to state-of-art approaches \cite{Lin:2009:FSS:1561638.1561679,DBLP:journals/jiis/0001KL12} which utilize constant functions to express local patterns, relies on offering an expressive technique to represent patterns as polynomial of arbitrary degrees. Furthermore, we present a fitting algorithm which can compute the polynomial coefficients for a sliding window segment in linear time, therefore our method offers superior expressiveness without compromising run-time complexity.

Empirical results conducted on datasets from the health care domain, demonstrate qualitative improvements compared to the state-of-art techniques in terms of classification accuracy, as is detailed in Section~\ref{resultsSec}. We experimented with the datasets from \cite{DBLP:journals/jiis/0001KL12}, and in order to widen the variety of data we introduce three new datasets. Our method wins in all four datasets with a clear statistically significance in three of them, proving the validity of our method. We add experiments regarding the running time of the method and we show that our method is practically fast and feasible.

% An example of a floating figure using the graphicx package.
% Note that \label must occur AFTER (or within) \caption.
% For figures, \caption should occur after the \includegraphics.
% Note that IEEEtran v1.7 and later has special internal code that
% is designed to preserve the operation of \label within \caption
% even when the captionsoff option is in effect. However, because
% of issues like this, it may be the safest practice to put all your
% \label just after \caption rather than within \caption{}.
%
% Reminder: the "draftcls" or "draftclsnofoot", not "draft", class
% option should be used if it is desired that the figures are to be
% displayed while in draft mode.
%

\section{Related Work}

\subsection{Time-Series Representations}

In order to understand the regularities embedded inside time-series, a large number of researchers have invested efforts into deriving and discovering time series representations. The ultimate target of representation methods is to encapsulate the regularities of time-series patterns by omitting the intrinsic noise. Discrete Fourier transforms have attempted to represent repeating series structures as a sum of sinusoidal signals \cite{Faloutsos:1994:FSM:191839.191925}. Similarly, wavelet transformations approximate a time-series via orthonormal representations in the form of wavelets \cite{KinPongChan1999}. However, such representations perform best under the assumption that series contain frequently repeating regularities and little noise which is not strictly the case in real-life applications. Singular Value Decomposition is a dimensionality reduction technique which has also been applied to extract latent dimensionality information of a series \cite{Cadzow83}, while supervised decomposition techniques have aimed at incorporating class information into the low-rank data learning \cite{DBLP:conf/aaai/GrabockaNS12}. 

In addition to those approaches, researchers have been also focused on preserving the original form of the time series without transforming them to different representations. Nevertheless, the large number of measurement points negatively influence the run-time of algorithms. Attempts to shorten time series by preserving their structure started by linearly averaging chunks of series points. Those chunks are converted to a single mean value and the concatenation of means create a short form known as a Piecewise Constant Approximation \cite{DBLP:journals/kais/KeoghCPM01}. A more sophisticated technique operates by converting the mean values into symbolic form into a method called Symbolic Aggregate Approximation, denoted shortly as SAX \cite{Lin:2007:ESN:1285960.1285965,Wei:2006:SEI:1193207.1193373}. SAX enables the conversion of time-series values into a sequence of symbols and offers the possibility to semantically interpret series segments. Further sophistication of lower bounding techniques have advanced the representation method towards efficient indexing and searching \cite{Shieh:2008:ISI:1401890.1401966}, enabling large scale mining of time series \cite{Camerra:2010:IIM:1933307.1934553}. Nonlinear approximations of the series segments have also been proposed. For instance least squares approximation of time series via orthogonal polynomials have been proposed for segmentation purposes in a hybrid sliding/growing window scenario \cite{Fuchs:2010:OST:1907655.1908058}. Throughout this paper we will propose a novel representation technique based on the utilization of polynomial functions of an arbitrary degree to approximate sliding windows of a time series. Our method brings novelty in converting the coefficients into literal representations, while the ultimate form is the frequency of the literal words constructed per each sliding window.

\subsection{Time-Series Similarity Metrics}

The time-series community has invested considerable efforts in understanding the notion of similarity among series. Time series patterns exhibit high degrees of intra and inter class variation, which is found in forms of noisy distortions, phase delays, frequency differences and signal scalings. Therefore, accurate metrics to evaluate the distance among two series play a crucial role in terms of clustering and classification accuracy. Euclidean distance, commonly known as the $L_2$ norm between vectors, is a fast metric which compares the offset of every pair of points from two series, belonging to the same time stamp index. Despite being a fast metric of linear run-time complexity, the Euclidean distance is not directly designed to detect pattern variations. A popular metric called Dynamic Time Warping (DTW) overcomes the deficiencies of the Euclidean distance by allowing the detection of relative time indexes belonging to similar series regions. DTW achieves highly competitive classification accuracies and is regarded as a strong baseline \cite{Ding:2008:QMT:1454159.1454226}. Even though DTW is slow in the original formulation having a quadratic run-time complexity, still recent techniques involving early pruning and lower bounding have utilized DTW for fast large scale search \cite{Rakthanmanon2012}. 

Other techniques have put emphasis on the need to apply edit distance penalties for assessing the similarity between time series \cite{Chen:2004:MLE:1316689.1316758, Chen:2005:RFS:1066157.1066213}. Such methods are inspired by the edit distance principle of strings which counts the number of atomic operations needed to convert a string to the other. In the context of time series the analogy is extended to the sum of necessary value changes needed for an alignment. Other approaches have put emphasis on detecting the longest common subsequence of series, believing in the assumption that time series have a fingerprint segment which is the most determinant with respect to classification \cite{Vlachos2002}. Detection of similarities in a streaming time-series scenarios motivated attempts to handle scaling and shifting in the temporal and amplitude aspects\cite{Chen2007}.

\subsection{Time-Series Classification}

During the past decade, most time series classification attempts have targeted the classification of short time series. Nevertheless, the definition of shortness might lead to ambiguous understandings, therefore we would dedicate some room for further clarifying our definition. For instance, assume a time-series dataset representing outer shapes of two different tree leaves, hence a binary classification. Such series include one single pattern each and do not exceed hundreds of data points, therefore we define them simply short. On the other hand assume we have an imaginary large dataset containing concatenated outer shapes of all the leaves of a forest. Then our imaginary task is to compare different forests for classifying which continent the forest is located in. In the second case, the time series are a combination of many local patterns occurring at varying frequencies and positions. This category of datasets is hereafter defined as long time series. 

\subsubsection{Classifying Short Time Series}

Classification of short time series has gathered considerable attraction in the literature. Among the initial pioneer methods and still one of the best performing ones is the nearest neighbor classifier accompanied by the DTW distance metrics, which constitute a hard-to-beat baseline \cite{Ding:2008:QMT:1454159.1454226}. Other powerful nonlinear classifiers like the Support Vector Machines have been tweaked to operate over time series, partially because originally the kernel functions are not designed for invariant pattern detection and partially because DTW is not a positive semi-definite kernel \cite{Gudmundsson2008}. Therefore the creation of positive semi-definite kernels like the Gaussian elastic metric kernel arose \cite{Zhang20102}. Another approach proposed to handle variations by inflating the training set and creating new distorted instances from the original ones \cite{DBLP:conf/pkdd/GrabockaNS12}.

\subsubsection{Classifying Long Time Series}

The classification of long time series focuses on long signals which are composed of one or more types of patterns appearing in unpredicted order and frequencies. Principally, the classification of those series has been mainly conducted by detecting the inner patterns and computing statistics over them. For instance, underlying series patterns have been expressed as the motifs and the difference between the motif frequencies has been utilized \cite{DBLP:conf/gfkl/BuzaS08}. Other approaches have explored the conversion of each sliding window segment into a literal word constructed by piecewise constant approximations and the SAX method \cite{Lin:2009:FSS:1561638.1561679,DBLP:journals/jiis/0001KL12}. The words belonging to each time series are gathered in a 'bag' and a histogram of the words is constructed. The histogram vectors are the new representations of the time series. Such a technique has been shown to be rotation-invariant, because the occurrence of a pattern is not related to its position \cite{DBLP:journals/jiis/0001KL12}. In contrast to the existing work, our novel study introduces an expressive histogram formulation based on literal words build from local pattern detection via polynomial approximations. Our model ensures scalability by computing in linear run-time.

%	
%	
%	
%	
%	
%	\begin{figure*}[!]
%	\centering
%	\includegraphics[scale=0.07]{sensitivity}
%	\caption{Hyper-parameter Search Sensitivity}
%	\label{polyFig}
%	\end{figure*}

\section{Proposed Method}

\begin{figure*}[!]
\centering
\includegraphics[scale=0.09]{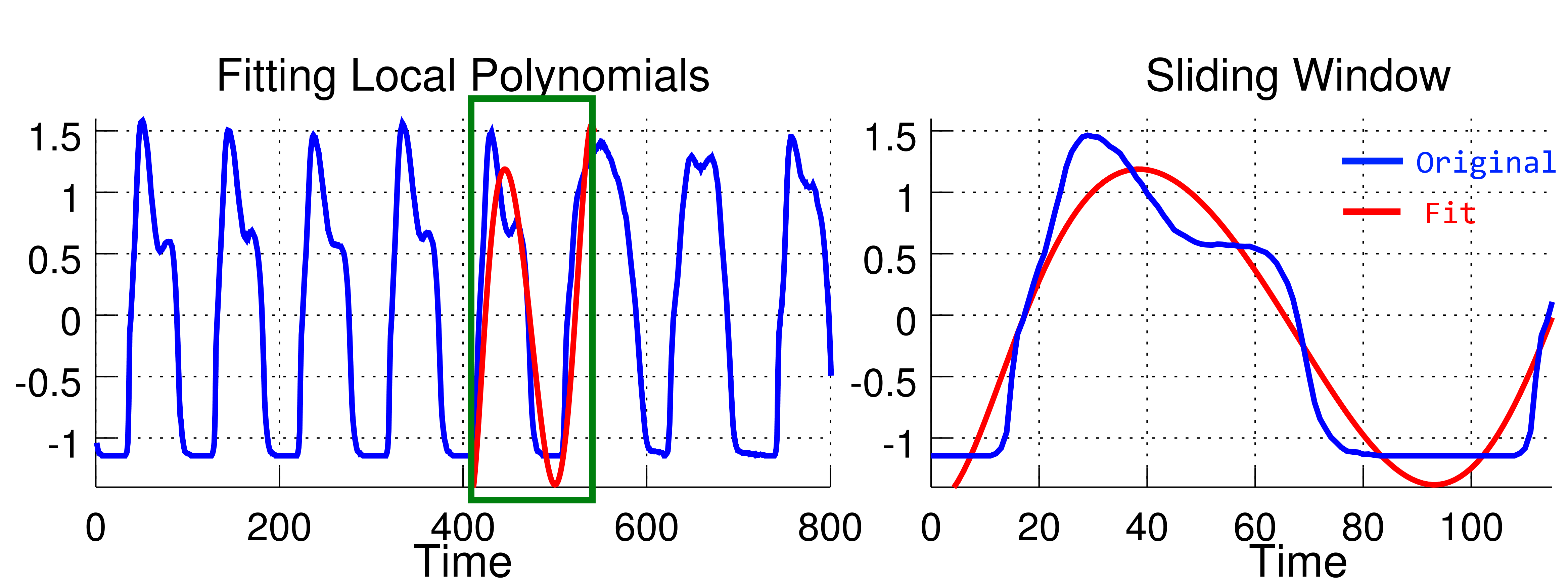}
\caption{Fitting a local polynomial of degree 8 to the sliding window region indicated by a green box. The plot on the right shows a scaled up version of the polynomial fit with coefficients $\beta =[8.4 \times 10^{-15}, -5.7 \times 10^{-12}, 1.6 \times 10^{-9}, -2.4 \times 10^{-7}, 2.15 \times 10^{-5}, -0.0011, 0.034, -0.36, -2.8]$ sorted from the highest monomial degree to the lowest. The time series on the left plot is a segment from the GAITPD dataset.}
\label{polyFig}
\end{figure*}

\subsection{Preamble Definitions}

\subsubsection{Alphabet} An alphabet is an ordered set of distinct symbols and is denoted by $\Sigma$. The number of symbols in an alphabet is called the size of the alphabet and denoted by $\alpha=|\Sigma|$. For illutration purposes we will utilize the Latin variant for the English language composed of the set of character symbols $\Sigma=\left( A,B,C,\dots,Y,Z \right)$.

\subsubsection{Word} A word $w \in \Sigma^{*}$ from an alphabet is defined as a sequence of symbols, therefore one sequence out of the set of possible sequences of arbitrary length $l$, defined as the Kleene star ${\Sigma^{*} := \cup_{l=0}^{\infty} \Sigma^{l}}$. For instance $\mbox{CACB}$ is a word from the English alphabet having length four.

\subsubsection{Polynomial} A polynomial of degree $d$ having coefficients $\beta \in \R^{d+1}$, is defined as a sum of terms known as monomials. Each monomial is a multiplication of a coefficient times the a power of the predictor value $X \in \R^{N}$, as shown in Equation~\ref{polyDefEq}. The polynomial can also be written as a linear dot product in case we introduce a new predictor variable $Z \in \R^{N \times (d+1)}$ which is composed of all the powers of the original predictor variable $X$.

\begin{eqnarray}
\label{polyDefEq}
\hat Y = \sum_{j=0}^d \beta_j X^j &=& Z \beta ,  \\ \nonumber 
\mbox{ where }  Z &:=& [ X^0 , X^1 , X^2, \dots, X^d ]
\end{eqnarray}

\subsubsection{Time Series} A time series of length $N$ is an ordered sequence of numerical values and denoted by $S \in \R^{N}$. The special characteristics of time-series data compared to plain vector instances is the high degree of correlation that close-by values have in the sequence. A time-series dataset containing $M$ instances is denoted as $T \in \R^{M \times N}$, assuming time series of a dataset have the same length.

\subsubsection{Sliding Window Segment} A sliding window segment is an uninterrupted subsequence of $S$ having length $n$ denoted by $S_{t,n} \in \R^{n}$. The time index $t$ represents the starting point of the series, while the index $n$ the length of the sliding window, i.e. $S_{t,n} = [ S_t,S_{t+1},S_{t+2},\dots,S_{t+n-1} ] $. The total number of sliding window segments of size $n$ for a series of length $N$ is $N-n$, in case we slide the window by incrementing the start index $t$ by one index at a time.

\subsection{Proposed Principle}

The principle proposed in this study is to detect local patterns in a time series via computing local polynomials. The polynomials offer a superior mean to detect local patterns compared to constant or linear models, because they can perceive information like the curvature of a sub-series. Furthermore, in case of reasonably sized sliding windows the polynomials can approximate the underlying series segment without over-fitting. In this paper, we demonstrate that the polynomial fitting for the sliding window scenario can be computed in linear run-time. Once the local polynomials are computed, we propose a novel way to utilize the polynomial coefficients for computing the frequencies of the patterns. The polynomial coefficients are converted to alphabet words via an equivolume discretization approach. Such a conversion from real valued coefficients to short symbolic words allows for the translation of similar polynomials to the same word, therefore similar patterns can be detected. We call such words symbolic polynomials. The words belonging to the time series are collected in a large 'bag' of words, (implemented as a list), then a histogram is created by summing up the frequency of occurrence for each words. Each row of a histogram encapsulates the word frequencies of time series, (i.e. frequencies of local patterns). A histogram row is the new representation of the time series and is used as a vector instance for classification.

\subsection{Local Polynomial Fitting}
\label{fitpolySec}

Our method operates by sliding a window throughout a time-series and computing the polynomial coefficients in that sliding window segment. The segment of time series inside the sliding window is normalized before being approximated to a mean 0 and deviation of 1. The incremental step for sliding a window is one, so that every subsequence is scanned. Computing the coefficients of a polynomial regression is conducted by minimizing the least squares error between the polynomial estimate and the true values of the sub-series. The objective function is denoted by L and is shown in Equation~\ref{lsDefEq}. The task is to fit a polynomial to approximate the real values $Y$ of the time series window of length $n$, previously denoted as $S_{t,n}$.
\begin{eqnarray}
\label{lsDefEq}
L(Y,\hat Y) &=& || Y - Z \beta ||^2 \\ \nonumber
 	Y &:=& [ S_t, S_{t+1}, S_{t+2}, \dots S_{t+n-1} ]
\end{eqnarray}

Initially, the predictors are the time indexes ${X=[0,1,\dots,n-1]}$ and they are converted to the linear regression form by introducing a variable $Z \in \R^{n \times (d+1)}$ as shown below in Equation~\ref{zEq}.

\begin{eqnarray}
\label{zEq}
Z = \left( \begin{array}{c c c c}
0^0 & 0^1  & \dots & 0^{d}  \\
1^0 & 1^1 & \dots & 1^{d} \\
\vdots & \vdots & \dots & \vdots \\
{(n-2)}^0 & {(n-2)}^1  & \dots & {(n-2)}^d  \\
{(n-1)}^0 & {(n-1)}^1 & \dots & {(n-1)}^{d}
\end{array} \right) 
\end{eqnarray}

The solution of the least square system is conducted by solving the first derivative with respect the polynomial coefficients $\beta$ as presented in Equation~\ref{lsSolEq}. 

\begin{eqnarray}
\label{lsSolEq}
\frac{ \partial L(Y,\hat Y)}{\partial \beta} = 0 &\mbox{ leads to }& \beta = \left( Z^{T}Z \right)^{-1} Z^{T} Y
\end{eqnarray}

A typical solution of a a polynomial fitting is provided in Figure~\ref{polyFig}. On the left plot we see an instantiation of a sliding window fitting. The sliding window of size $120$ is shown in the left plot, while the fitting of the segment inside the sliding window segment is scaled up on the right plot. Please note that inside the sliding window the time is set relative to the sliding window frame from $0$ to $119$. The series of Figure~\ref{polyFig} is a segment from the GAITPD dataset.

Since the relative time inside each sliding window is between $0$ and $n-1$, then the predictors $Z$ are the same for all the sliding windows of all time series. Consequently, we can pre-compute the term $P=\left( Z^{T}Z \right)^{-1} Z^{T}$ in the beginning of the program and use the projection matrix $P$ to compute the polynomial coefficients $\beta$ of the local segment $Y$ as $\beta=P Y$. Algorithm~\ref{fitAlg} describes the steps needed to compute all the polynomial coefficients of the sliding windows (starting at $t$) of every time series (indexed by $i$) in the dataset. For every time series we collect all the polynomial coefficients in a bag, denoted as $\Phi^{(i)}$. The outcome of the fitting process are the bags of all time series $\Phi$. Please note that the complexity of fitting a polynomial to a sliding window is linear and the overall algorithm has a complexity of $O(M \cdot d \cdot n \cdot N)$, which considering $d << N, d<<M$ and $n<<N$, means linear run-time complexity in terms of $N$ and $M$, that is $O(M \cdot N)$.

\begin{algorithm}[h]
\caption{Polynomial Fitting of a Time-Series Dataset}
\label{fitAlg}
\begin{algorithmic}[1]
\REQUIRE Dataset $T \in \R^{M \times N}$, Sliding window size $n$, Polynomial degree $d$
\ENSURE $\Phi \in \R^{M \times (N-n) \times (d+1)}$

\STATE $P \leftarrow \left( Z^{T} Z \right)^{-1} Z^{T}$
 \FOR{ $i \in \{ 1 \dots M \}$ }
	\STATE $\Phi^{(i)} \leftarrow \emptyset$
 \FOR{$t \in \{ 1 \dots N-n \}$ }
	\STATE $Y \leftarrow [ S_t^{(i)},S_{t+1}^{(i)},\dots S_{t+n-1}^{(i)} ]$
	\STATE $\beta \leftarrow P Y$
	\STATE $\Phi^{(i)} \leftarrow \Phi^{(i)} \cup \{ \beta \} $
 \ENDFOR
 \ENDFOR
%\RETURN $\left\( \Phi^{(i)} \right\)_{i=1\dots M}  $
\RETURN $( \Phi^{(i)} )_{i=1, \dots, M }  $
\end{algorithmic}
\end{algorithm}

\subsection{Converting Coefficients To Symbolic Words}
\label{wordsSec}

\begin{figure*}[t]
\centering
\includegraphics[scale=0.35]{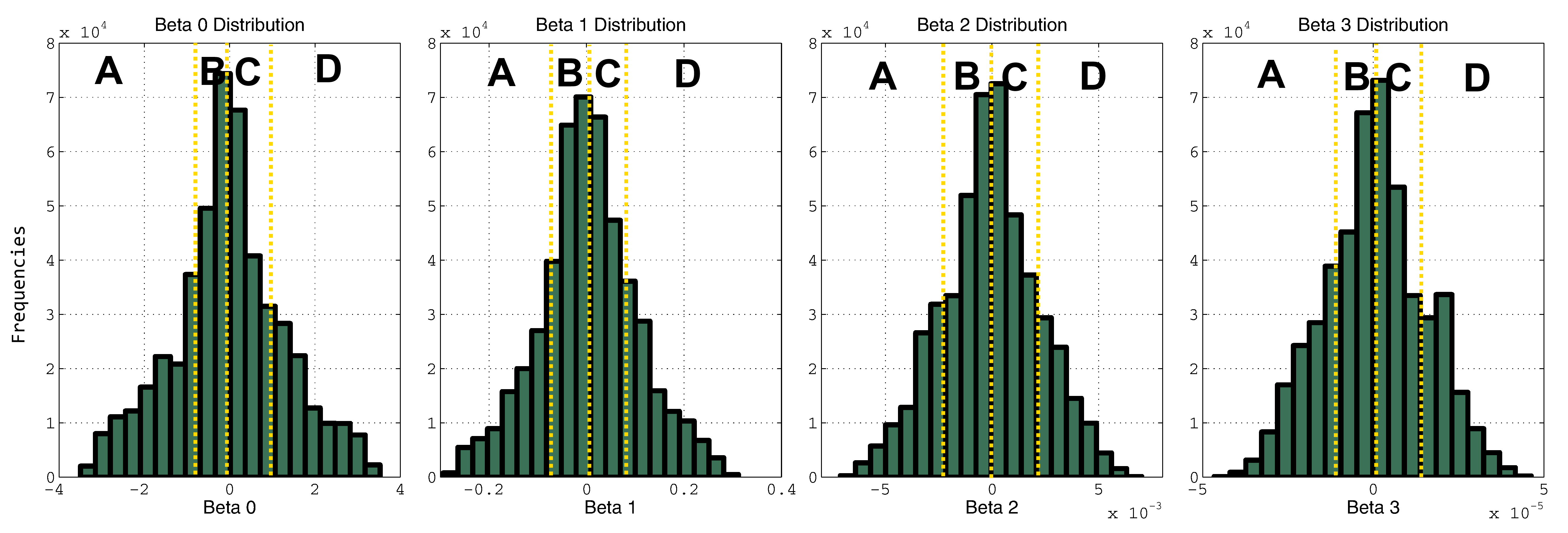}
\caption{Equivolume Discretization of Polynomial Coefficients. The illustration depicts the histogram distributions of a third degree polynomial fit over the sliding windows of the RATBP dataset. Each plot shows the values of a polynomial coefficient versus the frequencies. The alphabet size is four corresponding to the set $\{A,B,C,D\}$. The quantile threshold points are shown by dashed yellow lines.}
\label{coeffsFig}
\end{figure*}

The next step of our study is to convert the computed polynomial coefficients $\Phi$ from Algorithm~\ref{fitAlg} into words. The principle of conversion is to transform each of the $d+1$ coefficient of every $\beta$ of $\Phi$ to one symbol. Therefore, the extracted words have lengths of $d+1$ symbols. For each of the $\beta$ values of the polynomial coefficients we construct the histogram distribution and divide it into regions of equal volume as shown in Figure~\ref{coeffsFig}. In the image we have divided the histogram into as many regions as the alphabet size ($\alpha=4$) we would like to utilize. Such a process is called an equivolume discretization. The thresholds between the regions are named quantile points and are defined in the figure as yellow lines. Dividing the histogram into $\alpha$ many regions is equivalent to sorting the coefficient values and choosing the threshold values corresponding to indexes multiple of $\frac{1}{\alpha}$. For instance, dividing the histogram into 4 regions for an alphabet of size 4 requires thresholds values corresponding to indexes at $\frac{1}{4},\frac{2}{4},\frac{3}{4}$ of the total number of values, which means that the each region has $25\%$ of the values. Formally, let us define a sorted list of the j-th coefficient values regarding all window segments as ${B^j \leftarrow \mbox{sort} \left( \left\{ \beta_j \; | \; \beta \in \Phi^{(i)}, i = 1,\dots,M  \right\} \right) }$ and let the size of this sorted list be $s^j \leftarrow |B^j|$. Then the $(\alpha-1)$ many threshold values are defined as $\mu^j_k \leftarrow B^j_{ \left \lfloor s^j \frac{k}{\alpha} \right \rfloor }, \forall k \in \{1,\dots \alpha-1\}$ and $\mu^j_{\alpha} \leftarrow \infty$.

\begin{algorithm}[h]
\caption{Convert Polynomial Coefficients to Words}
\label{wordAlg}
\begin{algorithmic}[1]
\REQUIRE Polynomial Coefficients $\Phi$, Alphabet Size $\alpha$
\ENSURE $W \in \R^{M \times (N-n) \times (d+1)}$
\STATE \COMMENT{Compute the thresholds}
 \FOR{$j \in \{ 0 \dots d \}$ }
 \STATE ${B^j \leftarrow \mbox{sort} \left( \left\{ \beta_j \; | \; \beta \in \Phi^{(i)}, i = 1,\dots,M  \right\} \right) }$ 
 \STATE $s^j \leftarrow |B^j|$
 \STATE $\mu^j_{\alpha} \leftarrow \infty$
\FOR{$k \in \{ 1 \dots \alpha - 1 \}$ }
 \STATE $\mu^j_k \leftarrow B^j_{ \left \lfloor s^j \frac{k}{\alpha} \right \rfloor }$
 \ENDFOR
 \ENDFOR
\STATE \COMMENT{Convert the coefficients to words}
\STATE $\Sigma \leftarrow \left\{ A,B,\dots,Y,Z \right\}$
 \FOR{ $i \in \{ 1 \dots M \}$ }
	\STATE $W^{(i)} \leftarrow \emptyset$
 \FOR{$\beta \in \Phi^{(i)}$ }
 \STATE $w \leftarrow \emptyset$
 \FOR{$j \in \{ 0 \dots d \}$ }
%	\STATE $c \leftarrow \Sigma_{\alpha}$
% \FOR{$k \in \{ 1 \dots \alpha - 1 \}$ }
%	\IF{$\beta_j < \mu_k^j$}
%	\STATE $c \leftarrow \Sigma_k$
%	\STATE {\bf break}
%	\ENDIF
% \ENDFOR
 \STATE $k \leftarrow \argmax_{k \in \{1,...,\alpha \} } \beta_j < \mu^j_k $
 \STATE $w \leftarrow w \circ \Sigma_k$
 \ENDFOR
\STATE $W^{(i)} \leftarrow W^{(i)} \cup \{ w \} $
 \ENDFOR
 \ENDFOR
\RETURN $( W^{(i)} )_{i=1, \dots, M }$
\end{algorithmic}
\end{algorithm}

Algorithm~\ref{wordAlg} describes the conversion of polynomial coefficients to symbolic form, i.e. words. The first phase computes the threshold values $\mu_k^j$ to discretize the distribution of each coefficient in an equivolume fashion. The second phase processes all the coefficients $\beta$ of time series sliding windows and converts each individual coefficient to a character $c$,  depending on the position of the $\beta$ values with respect to the threshold values. The concatenation operator is denoted by the symbol $\circ$. The characters are concatenated into words $w$ and stored in bags of words $W$. The complexity of this algorithm is also linear in terms of $N$ and $M$. In case a linear search is used for finding the symbol index $k$, then the complexity is $O(M \cdot N \cdot \alpha)$, while a binary search reduces the complexity to $O(M \cdot N \cdot \log(\alpha))$. Yet, please note that in practice $\alpha << N$ and $\alpha << M$, therefore the complexity is translated to $O(M \cdot N)$.

\begin{figure}[!]
\centering
\includegraphics[scale=0.3]{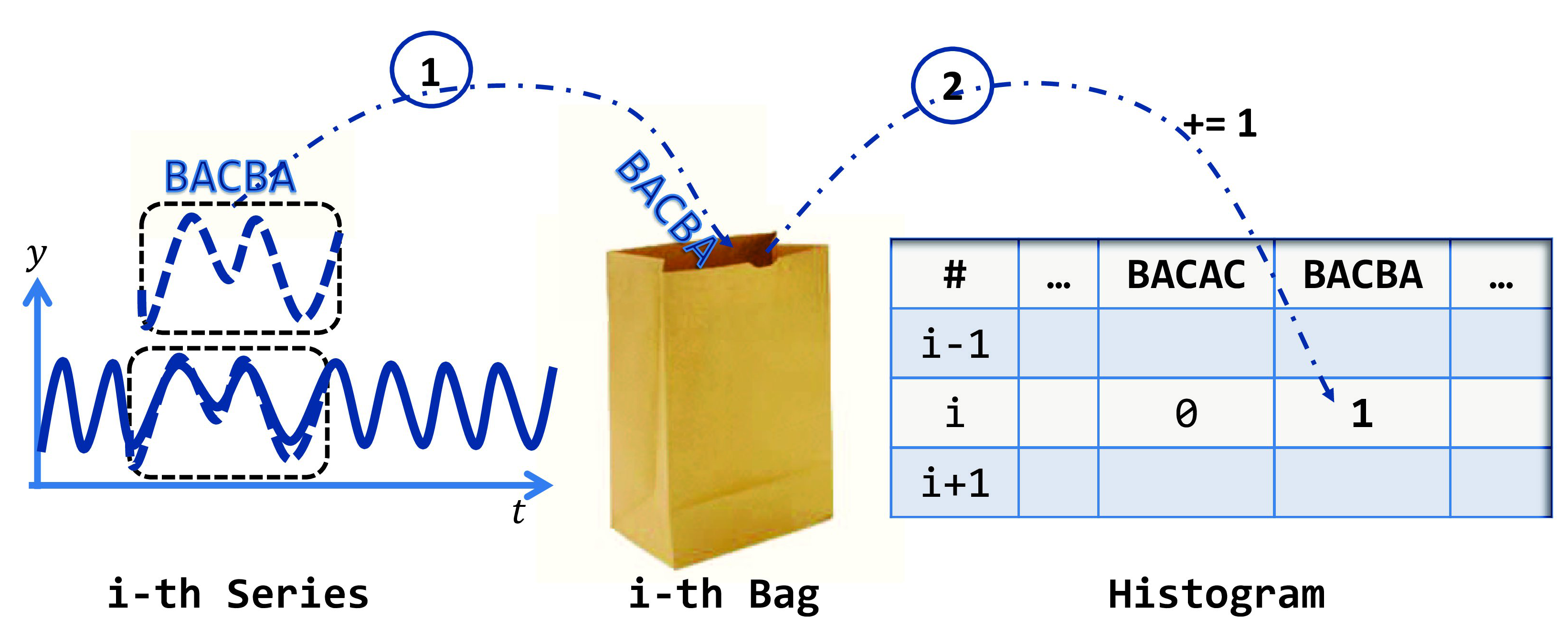}
\caption{Polynomial Words Collection and Histogram Population. In the first step all the words of a series are stored in one 'bag' per each time series. Then a histogram is initialized with a column of zeros for each word that occurs in any bag at least once. During the second step the word frequencies in the histogram are incremented upon each word found in a bag.}
\label{histFigure}
\end{figure}

\begin{figure*}[!]
\centering
\includegraphics[scale=0.13]{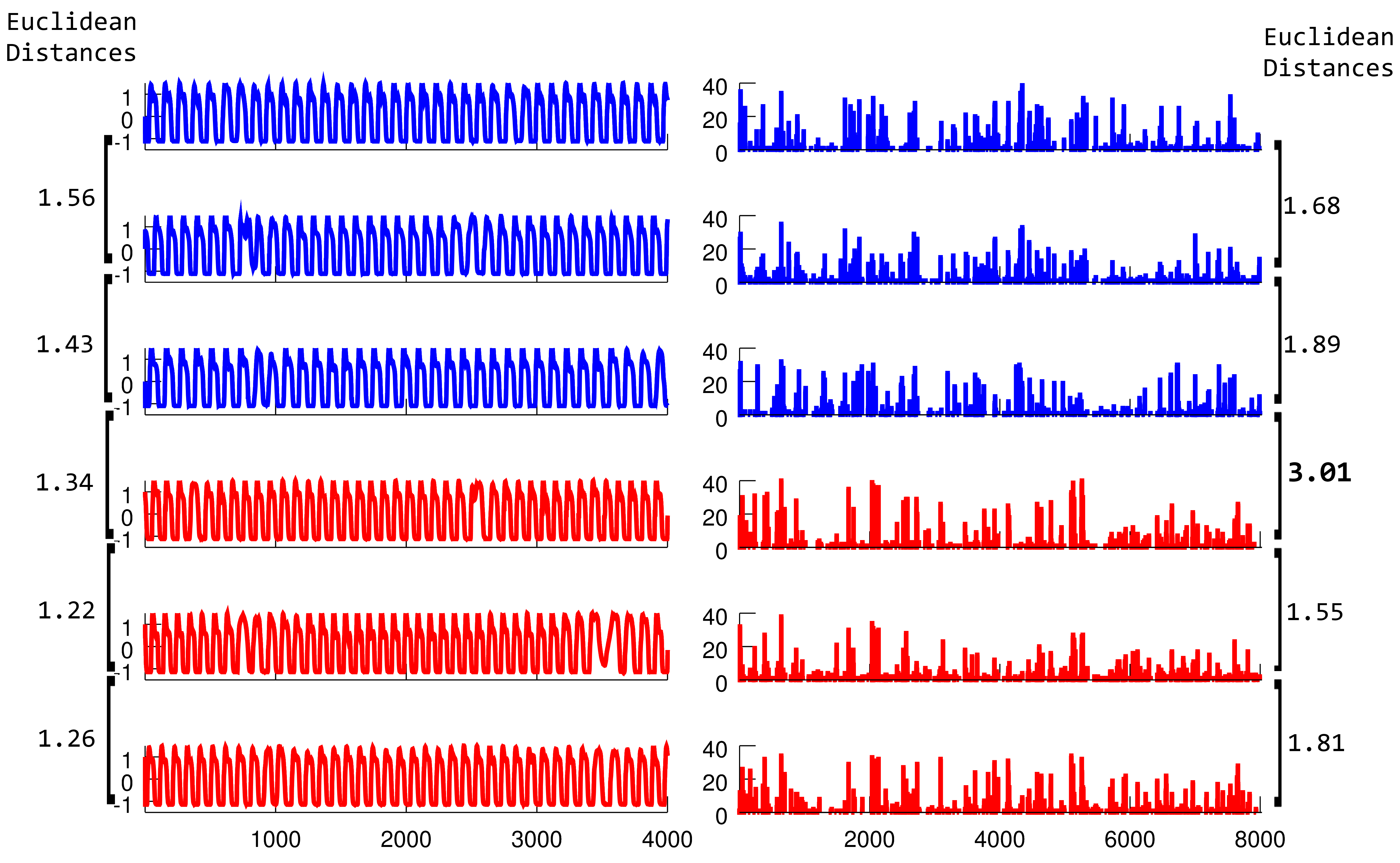}
\caption{GAITPD Dataset: Time series describing the gait in a few, randomly selected, control patients (blue) and Parkinson's disease patients (red). The original time series (x-axis is time and y-axis the normalized value) are displayed on the left column, while the respective histogram of each time series is shown on the right. The x-axis of the right plot represent 8026 words of the dictionary while the y-axis is the frequency of each word. The distances on the right show that histograms among a class have lower Euclidean distances, while the distance between the two histograms belonging to different classes is much higher, at a value of 3.01.}
\label{tsHist}
\end{figure*}

\subsection{Populating the Histogram}

Once we have converted our polynomial coefficients and converted them to words, the next step is to convert the words into a histogram of word frequencies, as depicted in Figure~\ref{histFigure}. The steps of the histogram population are clarified by Algorithm~\ref{histAlg}. The first step is to build a dictionary $D$, which is a set of each word that appears in any time series at least once. Then we create a histogram $H$ with as many rows as time-series and as many columns as there are words in the dictionary. The initial values of the histogram cells are 0. Each cell indicate a positive integer which semantically represent how many times does a word (column index) appear in a time series (row index). The algorithm iterates over all the words of a series and increases the frequency of occurrence of that word in the histogram.

\begin{algorithm}[h]
\caption{Populate the Histogram}
\label{histAlg}
\begin{algorithmic}[1]
\REQUIRE Word bags $W$
\ENSURE Histogram $H$
 \STATE \COMMENT{Build the dictionary}
 \STATE Ordered set dictionary $D \leftarrow \emptyset$
 \FOR{ $w \in W^{(i)}, \forall i \in \{ 1 \dots M \}$ }
 \IF{$ w \notin D $}
	\STATE $D \leftarrow D \cup \{ w \}$
 \ENDIF
 \ENDFOR
 \STATE \COMMENT{Build the histogram}
 \STATE $H \leftarrow \{0\}_{i=1,\dots,M \; j=1,\dots,|D|}$
 \FOR{ $w \in W^{(i)}, \forall i \in \{ 1 \dots M \}$ }
	\STATE Find $j$ with $D_j = w$ 
	\STATE $H_{i,j} \leftarrow H_{i,j} + 1$
 \ENDFOR
\RETURN $H$
\end{algorithmic}
\end{algorithm}

\begin{figure*}[!]
\centering
\includegraphics[scale=0.28]{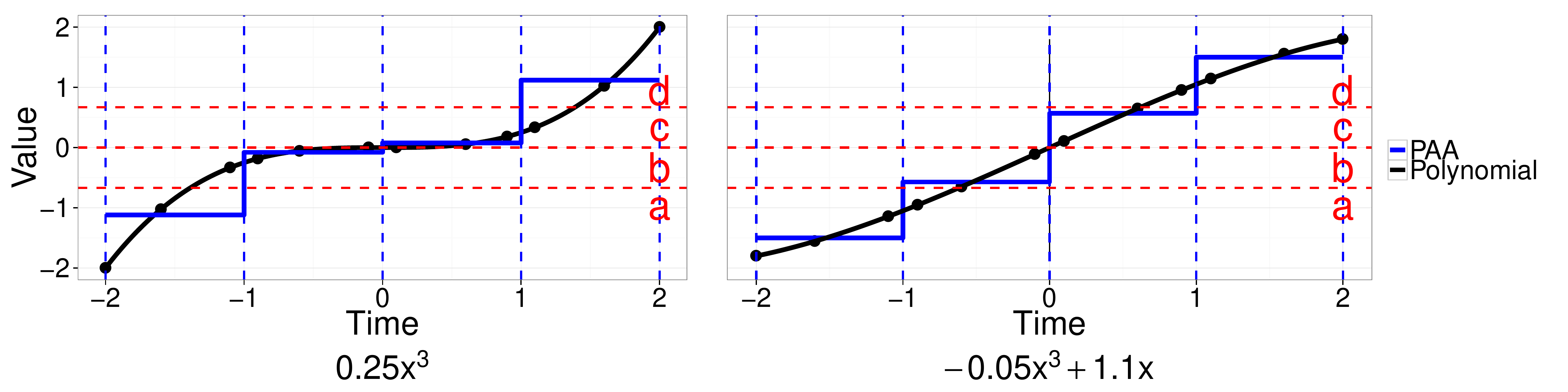}
\caption{Comparison of Symbolic Polynomial method to SAX. As can be seen a constant model named Piecewise Aggregate Approximation (PAA) fails to detect the curvature of the patterns and results in the same SAX word 'ABCD' for both series. }
\label{saxFig}
\end{figure*}

Once the histogram is populated, then each row of the histogram denotes a vector containing the frequencies of the dictionary words for the respective time series. Practically the row represent what local patterns (i.e. words) exist in a series and how often they appear. For instance Figure~\ref{tsHist} presents instances from the GAITPD dataset belonging to two types of patients in a binary classification task, healthy patients (blue) and Parkinson's Disease patients (red). In the left plot we show the original time series while on the right plot the histogram rows containing the polynomial words versus their frequencies. The parameters leading to the histogram for the GAITPD dataset are $n=100,\alpha=4,d=7$. As can be inspected the original time-series offer little direct opportunity to distinguish one class from the other and the series look alike. Moreover the Euclidean distance of adjacent series in the figure show that the Euclidean classifier would mistakenly classify the third instance of the blue class. In contrast the histograms are much more informative and it is possible to observe frequencies of local patterns which allow the discrimination of one class from the other. A complete distance matrix between blue $B$ and red $R$ instances is shown in Table~\ref{distancesTable}. As can be seen our histogram representations result in perfect accuracy in terms of nearest neighbor classification ({\bf bold}), while the original series result in 2 errors.

%\vspace{-0.2cm}

\begin{table}[!ht]
\centering
\caption{Distances: Time-series (left), Histogram (right)}
\resizebox{\linewidth}{!}{
\begin{tabular}{ll}
\begin{tabular}{| l | c | c | c | c | c | c |}
\hline
 &  $B_1$ & $B_2$ & $B_3$ & $R_1$ & $R_2$ & $R_3$ \\ \hline
$B_1$ & -  &  \cellcolor{blue!25} 1.6 &   \cellcolor{blue!25} \bf 1.28 & 1.4 & 1.29 & 1.4 \\ \hline
$B_2$ & \cellcolor{blue!25} 1.6   & - &  \cellcolor{blue!25} 1.4 & 1.5 & 1.4 & \bf 1.31 \\ \hline
$B_3$ & \cellcolor{blue!25} 1.28  & \cellcolor{blue!25} 1.4 &   - & 1.4  & \bf 1.27 & 1.4 \\ \hline
$R_1$ & 1.4   & 1.5 &  1.4 & - & \cellcolor{red!25} \bf 1.22 & \cellcolor{red!25} 1.4 \\ \hline
$R_2$ & 1.29  & 1.4 &   1.27 & \cellcolor{red!25} \bf 1.22  & - & \cellcolor{red!25} 1.26 \\ \hline
$R_3$ & 1.4   & 1.31 &  1.4 & \cellcolor{red!25} 1.4 & \cellcolor{red!25} \bf 1.26 & - \\ \hline
\end{tabular}
&
\begin{tabular}{| l | c | c | c | c | c | c |}
\hline
 &  $B_1$ & $B_2$ & $B_3$ & $R_1$ & $R_2$ & $R_3$ \\ \hline
$B_1$ & - & \cellcolor{blue!25} \bf 1.7 &  \cellcolor{blue!25} 2.3 & 2.9  & 2.6 & 2.5 \\ \hline
$B_2$ & \cellcolor{blue!25} \bf 1.7   & - &  \cellcolor{blue!25} 1.9 & 2.6  & 2.3 & 2.3 \\ \hline
$B_3$ & \cellcolor{blue!25} 2.3   & \cellcolor{blue!25} \bf 1.9 &   - & 3.0  & 2.6 & 2.8 \\ \hline
$R_1$ & 2.9   & 2.6 &   3.0 & - & \cellcolor{red!25} 1.6 & \cellcolor{red!25} \bf 1.4 \\ \hline
$R_2$ & 2.6  & 2.3 &   2.6 & \cellcolor{red!25} \bf 1.6  & - & \cellcolor{red!25} 1.8 \\ \hline
$R_3$ & 2.5   & 2.3 &  2.8 & \cellcolor{red!25} \bf 1.4 & \cellcolor{red!25} 1.8 & - \\ \hline
\end{tabular}
\end{tabular}
}
\label{distancesTable}
\end{table}

\subsection{Comparison To Other Methods}

The closest method comparable in nature to ours is the approach which builds histograms from SAX words \cite{Lin:2009:FSS:1561638.1561679,DBLP:journals/jiis/0001KL12}. However the SAX words are build from locally constant approximations which in general are less expressive than the polynomials of our approach. Figure~\ref{saxFig} demonstrate the deficiencies of the locally constant approximation in detecting the curvatures of a sliding window sub-series. In the experiment of Figure~\ref{saxFig}, we used an alphabet of size four and utilized the classical quantile threshold  for SAX, being values $\{-0.67,0,0.67\}$. Please note that the series are shown by black dots and represent normalized window segments. We have fitted both a constant model and our polynomial model to the series data. Assume we want to have a four character SAX word for each of the sliding windows segment. As can be easily seen the SAX words for both segments are 'ABCD'. On the other hand, referring to the coefficient values of Figure~\ref{coeffsFig}, we can see that the symbolic polynomial word belonging to polynomial $y(x)=0.25 x^3$ is 'CCBC', while the polynomial word belonging to $y(x)=-0.05x^3+1.1 x$ is 'BCDC'. As we can see our method can accurately distinguish the difference between those patterns, while the SAX method averages the content and looses information about their curvatures. We would like to point out clearly that our method is more expressive by needing the same complexity, in this case both methods use four characters. In terms of run-time, SAX needs only one pass through the data, so from the algorithmic complexity point of view both methods have same algorithmic complexity of $O(M \cdot N)$, i.e. number of series by their length. The complexity of our method has an multiplicative constant, which is the degree of the polynomial as shown in Algorithm~\ref{wordAlg}, however $d<<M,d<<N$. Finally, as will be shown in Section~\ref{resultsSec}, the classification results of our method are much better than SAX histograms.

\subsection{Classifier}

The classifier that we are going to use is the nearest neighbor method, which is a strong classifier in time-series classification \cite{Ding:2008:QMT:1454159.1454226} and is used by state-of-art methods \cite{DBLP:journals/jiis/0001KL12}. After converting the original time series into pattern frequency representations, in the form of rows of a histogram matrix, then each row will be treated as a vector instance. The nearest neighbor will utilize the Euclidean distance to compute the difference between histogram rows.

\section{Experimental Setup}

\subsection{Descriptions of Datasets}

\begin{table}[]
\centering
\caption{{\bf Statistics of Datasets} \newline }
\label{data_stat}
\begin{tabular}{|l|c|c|c|}
\hline
{\bf Dataset} & {\bf Number of Instances} & {\bf \mbox{Series Length}} & {\bf Number of Labels} \\ \hline  \hline
ECG2 & 250 & 2048 & 5 \\ \hline
GAITPD & 1552 & 4000 & 2 \\ \hline
RATBP & 180 & 2000 & 2 \\ \hline
NESFDB & 840 & 1800 & 2 \\ \hline
\end{tabular}
\label{dsStatTable}
\end{table}

All the experiments are based on datasets retrieved from Physionet, a repository of complex physiological signals primarily from the health care domain \cite{PhysioNet}. Our first dataset, ECG2, represents time series from the domain of Electrocardiography belonging to five different sources \cite{DBLP:journals/jiis/0001KL12}. The other datasets shown below were not used before in the realm of time-series classification, so we processed and adopted them for a classification task. Being the first paper to introduce them for time-series classification, we are dedicating some lines to explain the preparation of the datasets.

\begin{itemize}
\item{GAITPD:} The dataset contains measures of walking patterns (aka gait) from 93 patients with idiopathic Parkinson's Disease and 73 healthy (control) patients \cite{hausdorff2007rhy}. Measurements of the force underneath each foot was recorded for each patient via 8 sensors. We selected the total force of either feet as two independent series. The recorded gait time series were divided into equilength segments of 4000 measurement points. Finally we created a binary classification dataset by separating healthy and Parkinson's patients.
\item{RATBP:} Hypertension caused by salt consumption was tested over two types of rats, the Dahl salt sensitive (SS) rats and the consomic SS.13BN rats \cite{Bugenhagen2010}. Both high-salt and low-salt diets were provided to a total of 15 rats throughout a certain period of time. The amount of salt in the diet was alternated for each rat and the blood pressure series of each rat were measured. We segmented the blood pressure time series into equilength chunks of 2000 point measurements and utilized a binary categorization scheme by separating low-salt series from high-salt series.
\item{NESFDB:} Postural sway refers to the oscillations of the body while standing. A group of 27 individuals of young and old age groups were voluntarily tested for postural sway behaviors \cite{priplata2003vib}. Vibrating elements were implanted in the shoes of each volunteer beneath the fore foot and heel. Simulations on the patients were conducted by emitting low-pass filtered vibrational noise. A reflective marker was placed on the shoulders of the individuals and the sway was converted from a video recording into a time series describing the displacement of the marker. We utilized the series derived from both front and lateral video recordings as two instances. The time series were divided into two groups of sways, sway measurements with vibrational stimulus and measurements without stimulus.
\end{itemize}

The statistics of each dataset in terms of the number of instances, the length of each time series and the number of classes is summarized in Table~\ref{dsStatTable}. Please note that all the instances within one dataset have the same length.

\subsection{Baselines}

Let us name our method as SymPol, meaning {\bf Sym}bolic {\bf Pol}ynomials and refer to our method with the abbreviation form in the remaining sections. In order to evaluate the performance of SymPol, we compare against the following three baselines. 

\subsubsection{BSAX} refers to the method of constructing bags of SAX words from time series through a sliding window approach. The words occurring in the bags are used to populate a histogram of frequencies \cite{DBLP:journals/jiis/0001KL12}. A nearest neighbor method is applied to classify the histogram instances by treating the histogram rows as the new time-series representation. Comparing against this classifier will give chance to understand the benefit of polynomial approximation compared to constant models and will provide evidences on the state-of-art quality of the results.
\subsubsection{ENN} is the classical nearest neighbor classifier with the Euclidean $L_2$ loss as the distance metric. It operates over the whole time series, without segmenting the series for local patterns. The comparison against the plain nearest neighbor will show whether the detection of local patterns has more advantage than comparing the whole long series.
\subsubsection{DTWNN} differs from the Euclidean nearest neighbor classifier in defining a new distance metric for the comparison of two time series and performs well in time-series classification \cite{Ding:2008:QMT:1454159.1454226}. Dynamic Time Warping (DTW) operates by creating a matrix with all the possible warping paths, (i.e. alignment of pairs of indexes from two series), and selects the warping alignment with the smallest overall possible distance. DTW compares a full series without segmentations similarly to the Euclidean version of the nearest neighbor. Such comparison will both identify the benefits of the segmentation and also the benefits of local polynomials against global warping alignments.

\begin{table*}[!ht]
\centering
\caption{\bf Error Rate Results } 
\begin{tabular}{|l!{\vrule width 1pt}C{1.2cm}|C{1.2cm}!{\vrule width 1pt}C{1.2cm}|C{1.2cm}!{\vrule width 1pt}C{1.2cm}|C{1.2cm}!{\vrule width 1pt}C{1.2cm}|C{1.2cm}!{\vrule width 1pt}}
\hline
\multirow{2}{*}{\bf Dataset} & \multicolumn{2}{c!{\vrule width 1pt}}{\bf SymPol } & \multicolumn{2}{c!{\vrule width 1pt}}{\bf BSAX}  & \multicolumn{2}{c!{\vrule width 1pt}}{\bf ENN } & \multicolumn{2}{c|}{\bf DTWNN } \\ \cline{2-9}
 & \multicolumn{1}{r|}{ $\mu$ (mean) } & \multicolumn{1}{c!{\vrule width 1pt}}{ $\sigma$ (st.dev.) } & \multicolumn{1}{c|}{ $\mu$ (mean)} & \multicolumn{1}{c!{\vrule width 1pt}}{$\sigma$ (st.dev.) } & \multicolumn{1}{c|}{ $\mu$ (mean)} & \multicolumn{1}{c!{\vrule width 1pt}}{$\sigma$ (st.dev.) } & \multicolumn{1}{c|}{ $\mu$ (mean) } & \multicolumn{1}{c|}{$\sigma$ (st.dev.) } \\ \hline \hline 
ECG2 & {\bf 0.0000} & 0.0000 & 0.0080 & 0.0098 & 0.5480 & 0.0240 & 0.2120 & 0.0160 \\ \hline
GAITPD & {\bf 0.0238} &	0.0083	& 0.0548	& 0.0120	& 0.3924	& 0.0211	& 0.2468 &	0.0206 \\ \hline
RATBP & {\bf 0.1333}	& 0.0272	& 0.1889	& 0.0111	& 0.4389	& 0.0272	& 0.3333	& 0.0994
 \\ \hline
NESFDB & {\bf 0.4310}	& 0.0212	& 0.4405	& 0.0395	& 0.4929	& 0.0208	& 0.5440	& 0.0356 \\ \hline
\end{tabular}
\label{accuracyTable}
\end{table*}

\subsection{Reproducibility}
\label{reporSec}

Two different type of experiments were conducted in our study. The first empirical evidence focuses on the accuracy of our method with respect to classification of time series. The second experiment will analyze the computational run time of the methods. All the experiments were computed in a {\bf five folds  cross-validation} experimental setup. The time-series instances of each dataset were divided into 5 sets. In a circular fashion (repeated five times) each different set was once selected as the testing set, while the remaining four were used for training. Among the four sets used for training, one of them was selected as a validation set and the remaining three left as training. As a summary, all the combination of parameters were evaluated on the validation set and learned on the three training set, while the parameter values giving the smallest errors on the validation were selected. Those parameter values were finally evaluated over the testing set (learning from the three training sets) to report the final error rate. 

A grid search mechanism was selected for searching the hyperparameter values. Our method SymPol requires the tuning of three parameters, the  size of the sliding window $n$, the size of the alphabet $\alpha$ and the degree of the polynomials $d$. The size of the sliding window was selected among the range of $n \in \{ 100, 200, 300, 400\}$, while the size of the alphabet was picked from $\alpha \in \{4, 6, 8\}$. Lastly the degree of the polynomial was picked to be one of $d \in \{ 1,2,3,4,5,6,7,8 \}$. 

Similarly, the baseline named BSAX also requires the fitting of three hyperparameters. The length of a SAX word, denoted $|w|$, was selected from the range of $\{2,3,4,5,6,7,8,9\}$, while the size of the alphabet was selected among the values $\{ 4, 6, 8 \}$. The size of the sliding window is selected from a range of $\{100,200,300,400\}$, however those values were rounded to fit the length of a sax word. For instance if the length of a Sax word is 3, then the size of the sliding window was rounded from 100 to 102 in order for the sliding window to be equally divisible into three chunks. The hyperparameter values found in our experiments are shown in Table~\ref{hpsTable}, with ranges of multiple values due to different parameter searches per each different validation set.

\vspace{-0.2cm}
\begin{table}[!ht]
\centering
\caption{\bf Hyperparameter Search Results } 
\begin{tabular}{| l | c | c | c | c | c | c |}
\hline
\multirow{2}{*}{\bf Dataset} &  \multicolumn{3}{c|}{\bf SymPol}  & \multicolumn{3}{c|}{\bf BSAX} \\ \cline{2-7} 
 & $n$ & $\alpha$ & $d$ & $n$ & $|w|$ & $\alpha$ \\ \hline \hline
ECG2 & 100 & 4 & 3,4 & 100 & 4,6 & 4,6 \\ \hline
GAITPD & 100 & 4,6 & 7,8 & 100 & 6,7 & 6,8  \\ \hline
RATBP & 100 & 4,6 & 4,5,6,7 & 100 & 4,6,7 & 4,6,8  \\ \hline
NESFDB & 100,200 & 4,6,8 & 3,4,5 & 100,200,300 & 3,4 & 4,6  \\ \hline
\end{tabular}
\label{hpsTable}
\end{table}

\begin{table*}[!ht]
\centering
\caption{\bf Run Time Results (seconds)} 
\begin{tabular}{|l!{\vrule width 1pt}C{1.2cm}|C{1.2cm}!{\vrule width 1pt}C{1.2cm}|C{1.2cm}!{\vrule width 1pt}C{1.2cm}|C{1.2cm}!{\vrule width 1pt}C{1.2cm}|C{1.2cm}!{\vrule width 1pt}}
\hline
\multirow{2}{*}{\bf Dataset} & \multicolumn{2}{c!{\vrule width 1pt}}{\bf SymPol } & \multicolumn{2}{c!{\vrule width 1pt}}{\bf BSAX}  & \multicolumn{2}{c!{\vrule width 1pt}}{\bf ENN } & \multicolumn{2}{c|}{\bf DTWNN } \\ \cline{2-9}
 & \multicolumn{1}{r|}{ $\mu$ (mean) } & \multicolumn{1}{c!{\vrule width 1pt}}{ $\sigma$ (st.dev.) } & \multicolumn{1}{c|}{ $\mu$ (mean)} & \multicolumn{1}{c!{\vrule width 1pt}}{$\sigma$ (st.dev.) } & \multicolumn{1}{c|}{ $\mu$ (mean)} & \multicolumn{1}{c!{\vrule width 1pt}}{$\sigma$ (st.dev.) } & \multicolumn{1}{c|}{ $\mu$ (mean) } & \multicolumn{1}{c|}{$\sigma$ (st.dev.) } \\ \hline \hline 
ECG2 & 4.1	& 0.1	& 3.3	& 3.8	& 2.0	& 0.2	& 1651.5	& 52.9 \\ \hline
GAITPD & 1124.0	& 807.4	& 535.0	& 198.2	& 91.7	& 0.9	& 337555.1	& 20015.2 \\ \hline
RATBP & 27.2	& 36.7	& 1.8	& 0.8	& 0.7	& 0.0	& 1124.1	& 17.2 \\ \hline
NESFDB & 18.3	& 2.9	& 10.2	& 2.8	& 12.0	& 1.6	& 19535.2	& 304.7 \\ \hline
\end{tabular}
\label{runTimeTable}
\end{table*}

\begin{figure*}[!]
\centering
\includegraphics[scale=0.07]{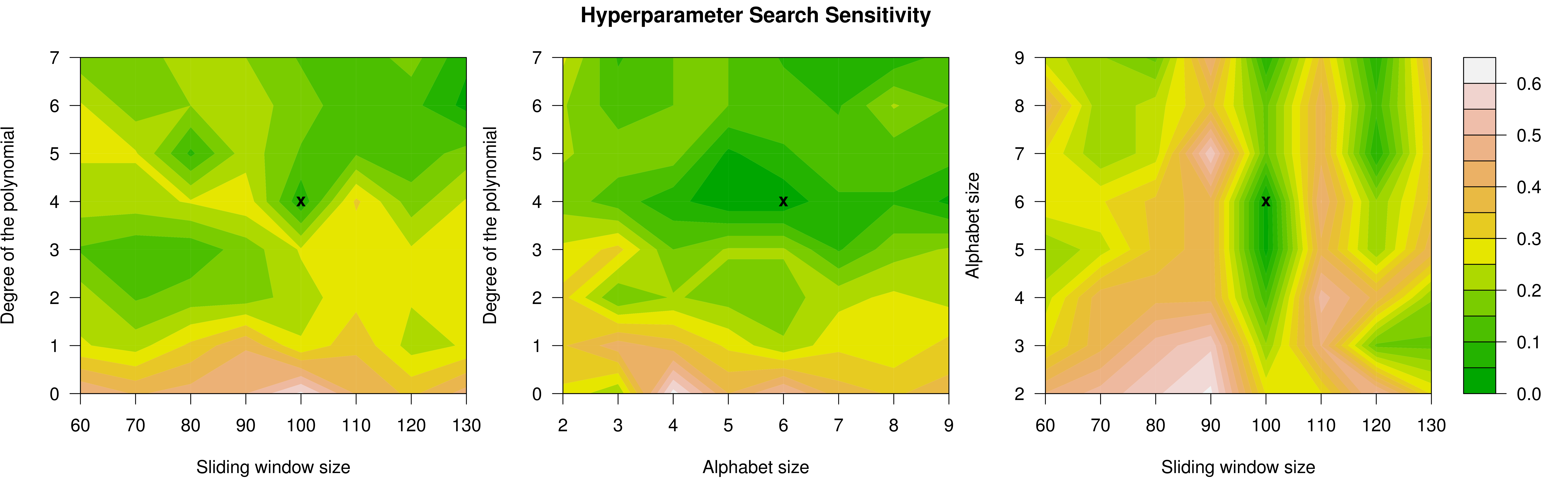}
\caption{Hyperparameter Search Sensitivity. Parameter search for one fold of the RATBP dataset resulting in the optimal values $n=100,\alpha=6,d=4$. In the two dimensional illustration the invisible parameter is fixed to the optimal.}
\label{sensFig}
\end{figure*}

\subsection{Results}
\label{resultsSec}

The classification accuracy results of our experiments are presented in Table~\ref{accuracyTable}. For our method SymPol and all the baselines we show the mean and the standard deviation of the five fold cross-validation experiments as described in the setup section.  The smallest error rate is highlighted in bold. 

As can be clearly seen our method demonstrates an unrivaled superiority. SymPol wins in all the four datasets and has a clear statistical significance in three of them ECG2, GAITPD, RATBP. Our method performs perfectly in the ECG2 dataset by having $100\%$ classification accuracy. In addition, SymPol reduces the error on the GAITPD dataset by $57\%$ with respect the closest baseline, while on the RATBP dataset the error is reduced by $29\%$.  

The second type of results represent the running times of the algorithms and is shown in Table~\ref{runTimeTable}. As can be clearly seen the Euclidean distance on the original dataset is the fastest method, which is a natural behavior because  no processing is done over series to extract histograms. The BSAX method is the next in terms of speed due to the computational advantage of the constant model which requires only one pass over the data. SymPol is positively positioned in terms of run time. As already analyzed before, the algorithmic complexity is comparable to the BSAX except for an additional constant, which is the polynomial degree. In datasets like ECG2, GAITPD and NESFDB the execution times are bigger by a small constant factor of two. The runtime constant in the RATBP dataset is higher because the hyperparameter search resulted to require a degree of 7. As a summary, we can clearly see that the method is practically very feasible and fast in terms of run time and is close even to techniques that require only a single scan over the time-series values.

\subsection{Hyper-parameter Search Sensitivity}

As presented in Section~\ref{reporSec} our hyperparameter search technique is  the grid search, where we scan for all the possible combination of one parameter's values to all the possible values of other parameters. As Figure~\ref{sensFig} shows, the error rate is nonlinear with respect to the parameter values of the method. Therefore, a grid search mechanism is practically suitable, because gradient based methods would have resulted in local optima while nonlinear optimization techniques would require much more computations than the grid. As can be seen in the plot, the grid search could successfully detect the global optimum in the region denoted by a mark.

\section{Conclusion}

In this study we presented a novel method to classify long time series, which are composed of local patterns. Local polynomial approximations are computed in a sliding window approach for each normalized segment under the sliding window. The computed polynomial coefficients are converted to symbolic forms (i.e. literal words) via an equivolume discretization procedure. Thresholds for the distribution of the values of each coefficient are determined to split the coefficient's histogram into equal regions and each region is assigned an alphabet symbol. In a second step all the polynomial coefficients are transformed into characters by locating them within the threshold values of the histogram and assigning the region symbol. The final literal representation of a polynomial is a word composed of the concatenation of each coefficient's character, in the order of the coefficient's monomial degrees. Once the bags of words are computed then a histogram is populated with the frequencies of each word in a time series. We presented a linear time technique to compute the polynomial approximation of a sliding window segment, while the overall method has a run time complexity which is linear in terms of the series points.

The classification accuracy of the nearest neighbor method utilizing the histogram rows that our method computed was compared against the performance of three baselines. Our method won all the experiments, most of them with a statistically significant margin. Furthermore, empirical results demonstrate that our method has a practically feasible running time performance, comparable even to the fastest methods which require a single scan over the time series.


\section*{Acknowledgment}

Blind Review



\bibliographystyle{IEEEtran}

\bibliography{allrelated}
%\begin{thebibliography}{1}
%\end{thebibliography}

\end{document}